\pdfoutput=1

\documentclass[11pt]{article}

\usepackage[preprint]{acl}

\usepackage{times}
\usepackage{latexsym}
\usepackage{adjustbox}
\usepackage{hyperref}

\usepackage[T1]{fontenc}

\usepackage[utf8]{inputenc}

\usepackage{microtype}

\usepackage{inconsolata}

\usepackage{graphicx}

\usepackage{microtype}
\usepackage{graphicx}
\usepackage{subfigure}
\usepackage{booktabs} 

\usepackage{amsmath}
\usepackage{amssymb}
\usepackage{mathtools}
\usepackage{amsthm}

\usepackage[capitalize,noabbrev]{cleveref}

\theoremstyle{plain}

\theoremstyle{definition}

\theoremstyle{remark}

\title{Can Large Language Models abstract Medical Coded Language?}

\author{Simon A. Lee \\
Department of Computational Medicine\\
University of California, Los Angeles\\
Los Angeles, California, USA 90095\\
  \texttt{simonlee711@g.ucla.edu} \\\And
  Timothy Lindsey \\
  Bioinformatic IDP\\
University of California, Los Angeles \\
Los Angeles, California, USA 90095\\
  \texttt{timlindsey32@g.ucla.edu} \\}

\begin{document}
\maketitle
\begin{abstract}
Large Language Models (LLMs) have become a pivotal research area, potentially making beneficial contributions in fields like healthcare where they can streamline automated billing and decision support. However, the frequent use of specialized coded languages like ICD-10, which are regularly updated and deviate from natural language formats, presents potential challenges for LLMs in creating accurate and meaningful latent representations. This raises concerns among healthcare professionals about potential inaccuracies or ``hallucinations" that could result in the direct impact of a patient. Therefore, this study evaluates whether large language models (LLMs) are aware of medical code ontologies and can accurately generate names from these codes. We assess the capabilities and limitations of both general and biomedical-specific generative models, such as GPT, LLaMA-2, and Meditron, focusing on their proficiency with domain-specific terminologies. While the results indicate that LLMs struggle with coded language, we offer insights on how to adapt these models to reason more effectively.

\end{abstract}

\section{Introduction}

The integration of Large Language Models (LLMs) into healthcare research represents a transformative shift in the analysis, interpretation, and utilization of medical data to enhance patient care. With their unparalleled ability to understand and generate human-like text, LLMs are increasingly used in diverse healthcare applications, including parsing \cite{goel2023llms} and summarizing clinical information \cite{van2023clinical} and providing clinical decision support \cite{lee2024emergency, cui2024llmsbased, jiang2023health}.

However, challenges persist within the complex healthcare ecosystem, particularly regarding the limitations of these technologies in clinical settings \citep{wornow2023shaky}. A primary concern is the ability of LLMs to accurately learn latent representations of coded language commonly used in healthcare, which includes various standardized codes such as International Classification of Diseases (ICD) codes, procedure codes (PCT), medication codes, and laboratory codes \citep{cimino1996coding, alexander2003overview}. These codes are organized into structured classification groups and play a critical role, but their regular updates and deviation from typical natural language complicate the ability of LLMs and more broadly machine learning models to develop meaningful and up-to-date representations \cite{cartwright2013icd}. Additionally, the scarcity of publicly available clinical notes poses a challenge for learning these subtle coded patterns due to their limited volume. These specific languages are often overshadowed by clinical abstracts and papers, which typically omit such codes. These documents are commonly used in pre-training efforts aimed at acquiring clinical vocabulary, as demonstrated in several foundation models \cite{alsentzer2019publicly, bolton2024biomedlm}.

 \begin{figure*}[t]
   \centering 
   \includegraphics[width=\textwidth]{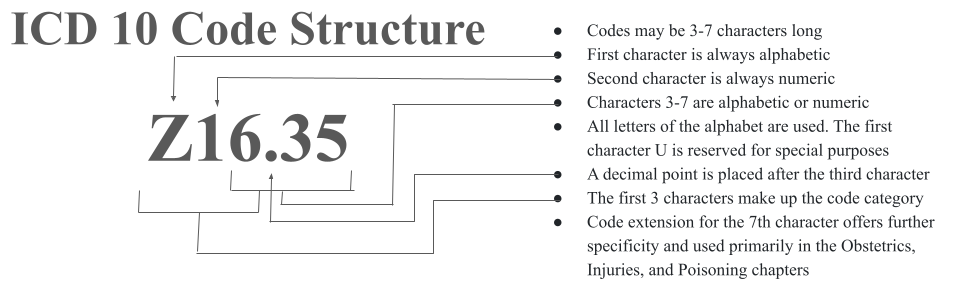} 
   \caption{Each Digit in a Medical Code corresponds to an encoded structural meaning}
   \label{fig111} 
 \end{figure*} 

While LLMs may struggle to adapt to recently updated codes, it is incorrect to suggest that these codes are comprised solely of arbitrary alphanumeric characters. The rule-based structure of medical coding systems ensures that each subsequent digit narrows the search for the correct label corresponding to a specific code, as demonstrated in Figure \ref{fig111}. Hence, this work evaluates the capabilities of LLMs in accurately generating names from medical codes. This potential deficiency could affect research groups focusing on summarization, automated billing, and inference-based tasks. It highlights the need for better representation of codes that do not conform to a natural language format, not only in healthcare but across various domains. Although previous research has explored numerical representations \citep{golkar2023xval}, to our knowledge, this study is the first to assess the proficiency of LLMs with medical codes.

\paragraph{The Significance of Medical Codes in Healthcare}

Medical codes are typically entered by healthcare providers into a patient's Electronic Health Record (EHR). These codes represent standardized protocols that often serve as translations of the notes taken by providers about a patient, and are utilized for various purposes. For instance, ICD codes, a specific type of diagnostic medical code, are frequently used by researchers for statistical analyses. Commercially ICD codes are also used for billing purposes based on medical diagnoses. The healthcare community views automated coding as one of the major potential use cases of AI to healthcare; however, in its current state, the utility seems limited \cite{soroush2024large}. Hence, inaccuracies in these codes are seen as problematic, as LLM technologies are known to produce errors or ``hallucinations''. Given the lack of evaluation of these technologies in handling medical codes, this issue represents an important research question within the fields of healthcare and AI.

\section{Related Work}
\subsection{Predicting medical codes from EHR}

Research in EHR has advanced through the application of computational techniques to predict and match ICD codes \citep{lita2008large, parr2018automated}, which are critical for various healthcare processes, including billing, epidemiology, and research. The evolution of this field has recently shifted towards employing natural language processing (NLP) techniques to bridge the gap between unstructured patient text and structured International Classification of Diseases (ICD) codes. Early efforts, as demonstrated by \citep{singaravelan2021predicting, atutxa2017machine, jay2019construction, mullenbach2018explainable, uzuner2010extracting, spasic2010medication, kavuluru2015empirical}, utilized rule-based algorithms and sophisticated NLP methods to automate the coding process, aiming to enhance both the accuracy and efficiency of healthcare documentation.

The advent of deep learning and pretrained language models such as BERT has  transformed the approach to predicting ICD codes, representing a considerable advancement in the capabilities of this field. Studies by \citep{samonte2018icd, chen2019automatic, huang2022plm, amin2019mlt} highlight the shift towards sophisticated models that leverage pre-trained language representations. These advancements have not only improved the precision of medical code prediction but also deepened the understanding of the complexities of medical language.

\subsection{Predicting Y Codes from X Codes}

Another popular avenue of research within EHR involves leveraging existing medical codes to predict other relevant medical codes, effectively utilizing one type of code as a predictor for another. This method has seen various implementations, such as using diagnostic codes to anticipate procedure codes, a relationship explored in the works by \citep{subotin2014system, haq2017intelligent}. Additionally, some research has explored combining multiple types of codes to enhance the prediction of disease codes, as demonstrated in the study by \citep{djennaoui2015improvement}.. 

 \begin{figure*}[t]
   \centering 
   \includegraphics[width=\textwidth]{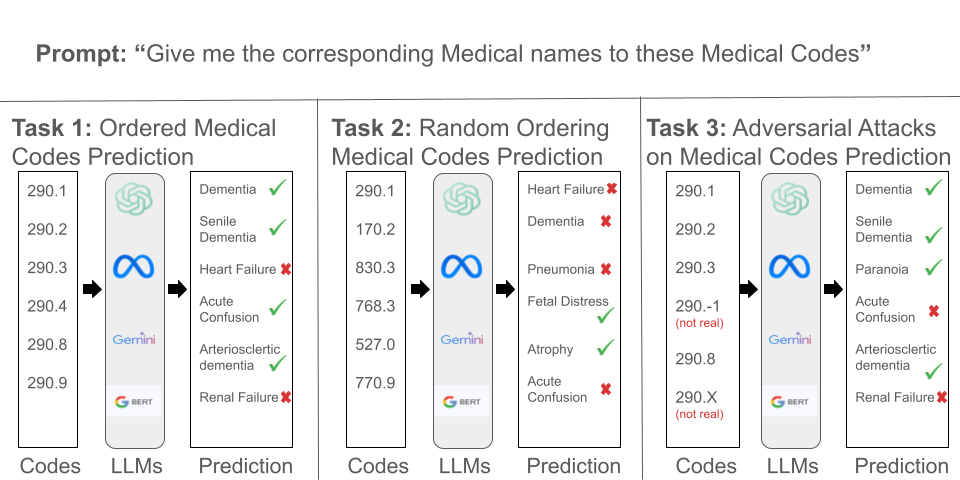} 
   \caption{An Overview of the Proposed Experiments to Evaluate the LLMs on Whether or Not They Can Properly Predict (Understand) the Corresponding Medical Codes.}
   \label{fig11} 
 \end{figure*} 

\subsection{LLMs and Healthcare}

LLMs have increasingly been recognized for their transformative potential in healthcare, as evidenced by recent research \citep{karabacak2023embracing}. One of the key utilities of LLMs lies in their capability to efficiently summarize clinical texts, a feature that has been highlighted in studies by \citep{van2023clinical, van2024adapted}. 

Another area where LLMs have shown considerable promise is in providing decision support. Research conducted by \citep{chen2024multimodal, belyaeva2023multimodal} demonstrates how LLMs can aid in constructing meaningful latent features in conjunction with multimodal data inputs, thereby facilitating more accurate predictions. 

Named-entity recognition (NER) represents yet another domain where LLMs have made significant advancements. The advent of BERT and its derivatives tailored for medical contexts—such as MedBERT \citep{liu2021med}, ClinicalBERT \citep{huang2019clinicalbert, mulyar2021mt}, and BioBERT \citep{lee2020biobert, shin2020biomegatron}—has led to a surge in models capable of identifying specific terms and concepts within clinical texts. These models have been instrumental in parsing and understanding complex medical documents, enhancing data extraction and processing tasks. 

More recently, the development of generative multimodal models such as MedPalm \citep{tu2024towards}, Clinical Camel \citep{toma2023clinical}, and Meditron \citep{chen2023meditron70b}, showcases ongoing innovation within the field, offering new avenues for utilizing generative LLMs in healthcare.

\section{Methods}
In this study, our objective is to determine whether LLMs can accurately generate labels from medical codes alone. We aim to assess the extent to which these models have developed the capability to understand the underlying structures and ontologies, and to generate text based on these coded languages. Additionally, we are actively working to minimize the generation of ``hallucinations'' by these models, as accuracy is important in the field of healthcare.


In our analysis, we utilize both commercial and open-source LLMs available on the \textit{Hugging Face} platform, as outlined in Figure \ref{fig11}. Additionally, we have restricted recent models like GPT-4 from accessing the internet during this assessment. Our aim is to investigate whether LLMs pretrained on biological, medical, or clinical data demonstrate a superior ability to learn and predict medical codes, potentially even with fewer parameters. The specific LLMs employed in our study are detailed in Table \ref{tab11}. Our selection criteria focused on a broad range of LLMs, varying in parameter size, domain expertise, and proprietary builds\footnote{The model parameters and training scheme of GPT-4 are largely unknown.}. For proprietary models, we utilize APIs to query and generate responses. For open-source models, we use Ollama as a platform to locally interact with LLMs. We use API's to interact with proprietary models.

\textit{We are aware of the continued development of new generative medical models. However, all analyses were conducted in January 2024, and we restricted our model selection to this timeframe.}

\begin{table}[h]
  \centering 
  \caption{Overview of various language models highlighting their source and parameter size.}
  \begin{adjustbox}{width=\columnwidth, center}
    \begin{tabular}{lcc}
    \toprule
      \textbf{Model} & \textbf{Parameters} & \textbf{Source} \\
      \midrule
      GPT-4 & N/A & Closed Source \\
      GPT-3.5 & N/A & Closed  Source \\
      Claude 3 & N/A & Closed Source \\
      Gemini & N/A & Closed Source \\
      LLaMA 2 & 70B & Open Source \\
      Mistral 7B & 7B & Open Source \\
      Meditron & 70B & Open Source \\
      Medllama2 & 7B & Open Source \\
      \bottomrule
    \end{tabular}
  \end{adjustbox}
  \label{tab11} 
\end{table}

\subsection{Experimental Design}
\label{exper}

We structured our experimental design into two specific tasks, each leading to a set of more granular downstream experiments. We anticipate that the complexity of these tasks will increase as the experiments progress further downstream.

\paragraph{Predicting Medical Chapters} Our initial experiment assesses whether Large Language Models (LLMs) can identify the categories or "chapters" corresponding to specific medical codes (e.g., ICD9: 001-139 corresponds to 'Infectious and Parasitic Diseases', ICD10 Procedure: 7 corresponds to 'Osteopathic Procedures'). If models can correctly make these distinctions, we hypothesize that LLMs either possess some understanding of the hierarchical structure or have memorized the medical code chapters.

\paragraph{Predicting Medical Codes} In our subsequent experiments, which evaluate the granularity of these models, we devised three distinct tests of varying difficulty levels. We included these levels because, in practice, any ICD code could be entered through a user query. Therefore, our aim is to assess and simulate potential real-life scenarios.

\begin{enumerate}
    \item \textbf{A List of Ordered Medical Codes:} Our initial experiment focused on predicting the medical conditions associated with specific ICD codes within the same "chapter" or category. This test evaluates the LLMs' ability not only to recognize the medical codes accurately but also to discern the underlying relationships reflected in their numerical structure. It explores how each subsequent digit in a medical code conveys specific information and adheres to a deliberate design. This experiment advances our initial hypothesis by suggesting that LLMs may understand the structural logic behind the medical codes, which could indicate the severity of the condition and pinpoint specific anatomical structures. Success in this test might reveal that LLMs either are aware of this inherent structure or have memorized the collection of codes.

    \item \textbf{A Randomly Ordered List of Medical Codes:} Our second experiment simulates a realistic scenario by testing the LLMs' ability to comprehend medical codes, mirroring real-world situations where a patient's clinical history might include lists of ICD codes from various chapters. We tasked each LLM with identifying the conditions associated with each medical code, using a random sample of indices for multiple tests. This test is challenging because the previous or subsequent codes may not provide contextual support for the medical condition in question. Success in this task would demonstrate a more comprehensive understanding.
    
    \item \textbf{Adversarial Attack:} Lastly, we designed an experiment to potentially exploit weaknesses in an LLM. Given that LLMs inherently respond to all queries unquestioningly, we introduced a test that intentionally incorporates malicious or fake medical codes among real ones. We created "fake ICD codes" that closely mimic real ICD codes (e.g., by altering the final digit) and then asked the LLM to generate labels for codes it presumed were genuine. This evaluation aimed to exploit potential hallucination behavior and to distinguish correct from incorrect codes. Successfully identifying these fake ICD codes would demonstrate a thorough understanding and awareness of medical coded language.
\end{enumerate}

\noindent We further illustrate these experiments by presenting an example query for each experiment in Table \ref{tab:models_overview}:

\begin{table*}[h!]
  \centering 
  \caption{An example query for each experimental design is provided. Our focus is on detecting "hallucinations," so we conduct several experiments that vary in difficulty from Experiment 1 to 3. \textcolor{red}{Red text} indicates the malicious examples that we intend to insert into our data for the adversarial attacks experiment.}
  \begin{tabular}{|p{0.3\textwidth}|p{0.6\textwidth}|}
  \toprule
    \textbf{Experiment} & \textbf{Queries} \\
    \midrule
    \textbf{1.} Ordered List & Predict the names of the diagnostics from this list of ICD10 Codes: [A12.01, A12.02, A12.03, A12.04, A12.08...]\\
    \textbf{2.} Randomly Ordered List & Predict the names of the diagnostics from this list of ICD10 Codes: [E570.1, G77.03, H81.06, A11.04, D43.01...] \\
    \textbf{3.} Adversarial Attack & Generate responses for only the real medical codes from this list of ICD10 Codes: [A12.01, A12.02, \textcolor{red}{A12.-03}, A12.04, \textcolor{red}{A12.0Z}...] \\
    \bottomrule
  \end{tabular}
  \label{tab:models_overview} 
\end{table*}

\subsection{Data}

Since medical codes are standardized across medical institutions, we utilize the open web to extract codes for the International Classification of Diseases (ICD-9/10), Current Procedural Terminology (CPT), medications (often referred to by the National Drug Code, NDC), and laboratory values (coded in LOINC). More details about the data sources are available in the appendix.

\subsection{Analysis}

We conduct all our tests and generate random samples using the Python programming language. To ensure uniform data inputs across all benchmarks, we set seeds to create consistent data partitions for each experiment. We employ Language Models from HuggingFace, using models locally on the \textit{ollama} platform (LLaMA2 70b, Medllama2, Mistral, and Meditron). Additionally, we utilize API keys from OpenAI (GPT-3.5/4), Anthropic (Claude-3), and Google (Gemini) to generate outputs from these models.

However, it is important to acknowledge the inherent variability in the outputs of LLMs. These models do not produce identical responses to user queries across different conversational sessions due to their stochastic decoding methods, which introduce randomness into the text generation process. Moreover, periodic model updates and the deliberate introduction of parameter randomization during inference contribute to this variability, leading to diverse outputs for the same inputs across different instances\footnote{\textit{This footnote details potential issues with obtaining exactly reproducible results.}}. This variability is a critical consideration for reproducing the results from our study. Thus, we wish to emphasize that all our studies were conducted in January 2024, and the models may have been updated since then.

\subsection{Model Evaluation}
For the various experiments conducted in our study, we have implemented distinct evaluation strategies to gain insights into the models' understanding and memorization of codes.

\paragraph{I. Predicting Medical Chapters}
In our first experiment, which involves predicting medical chapters, we assess accuracy based on how precisely the model generates the chapter name from its code. These generated responses are typically straightforward to validate due to their limited number of checks, so we employ human validators to confirm and measure this performance statistic.

\paragraph{II. Predicting Medical Franchise}
In our second experiment, we evaluate the similarity between the true text and the text generated by the model using BERT scores \citep{zhang2019bertscore}. We report the average BERT score for each LLM across all experiments in the study. Given the scalability of this metric with the number of experiments, we opted for a more robust statistic to measure the relative performance of the LLMs.

\paragraph{III. Adversarial Attack}
In the Adversarial Attack experiment, we modified the prompt to ask the LLM to generate predictions only for codes it identifies as real, while ignoring those considered fake. We assessed performance by checking whether the LLM ignored the fake codes or erroneously generated names for them, using a binary metric: a score of 0 indicates incorrect generation for a fake code, while a score of 1 indicates successful identification and ignoring of the fake code. A higher score is thus deemed better. We also asked the LLM to identify the fake codes in the query, using a binary metric to gauge performance. Human validators were employed to verify this accuracy and compile the results.

\section{Results}

\subsection{Generating Medical Chapter Names}

\begin{table}[h!]
  \centering 
  \caption{Accuracy of LLMs in Generating Correct Medical Chapters. The dagger (†) symbol indicates that the LLM failed to respond to the prompt as expected, despite various prompting attempts.}
  \begin{adjustbox}{width=\columnwidth}
    \begin{tabular}{lll}
      \toprule
      \textbf{Model} & \textbf{ICD9/10 Chapters} & \textbf{Procedure/Lab Chapters}\\
      \midrule
      Claude 3 & 1.0000 & 0.0588 \\ 
      Gemini & 0.8235 & 0.0000 \\ 
      GPT 3.5 & 1.0000 & 0.9412 \\ 
      GPT 4 & 1.0000 & 1.0000 \\ 
      LLaMA 2 & 0.3539 & 0.1765 \\ 
      Meditron & ---------$\dagger$ &  ---------$\dagger$ \\ 
      MedLLaMA2 & 1.0000 & 0.1176 \\ 
      Mistral & 1.0000 & 0.0588 \\ 
      \bottomrule
    \end{tabular}
  \end{adjustbox}
  \label{tab3} 
\end {table}

In Table \ref{tab3}, we present the results of our first experiment, which involved predicting chapter names. We observed a significant drop in performance when generating the correct names for Procedure and Lab Chapters compared to those for ICD 9/10 chapters. Consequently, early in our series of experiments, we decided to focus the LLM's tasks exclusively on predicting ICD 9/10 codes.

\subsection{Generating Medical Code Names}

 \begin{figure}[h!]
   \centering 
   \includegraphics[width=\columnwidth]{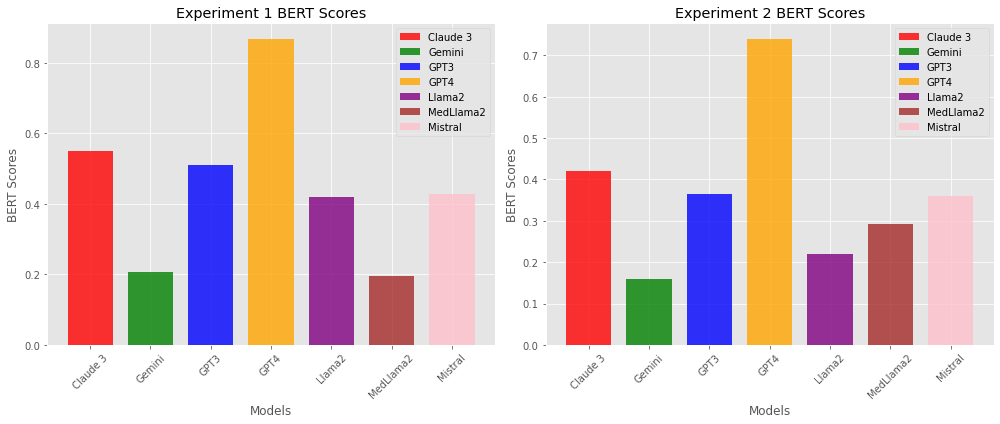} 
   \caption{The frequencies of ICD codes found in the publicly available MIMIC-IV Database and plotting their occurences to differentiate common from uncommon codes.}
   \label{bert} 
 \end{figure} 

In Figure \ref{bert}, we showcase the results from the medical codes study. We display the average BERT Score metrics for each LLM, effectively measuring the discrepancy between the generated label and the true label. We observed a significant performance disparity between the ordered (experiment 1) and unordered (experiment 2) setups, suggesting that LLMs struggle with the randomness of the codes they are tasked to generate. At the model level, GPT-4 appears to be the best-performing model, potentially indicating that the number of parameters may play a role in performance. This is consistent even in the reverse trajectory as indicated by the work of \cite{soroush2024large}.

\subsection{Adversarial Attack Experiment}

\begin{table}[h!]
  \centering 
  \caption{This table displays the results from the adversarial attack experiment. In the ``Correctly generating'' column, we see that all LLMs struggle to decipher between real and fake codes. This is evident because they all appear to erroneously generate labels for fake codes. GPT4 is a minor exception where they were able to differentiate a few of these fake codes. Also when we prompted the LLMs to select the fake codes from a mixture of codes, we see that all LLMs struggled to find and indentify the fake codes as indicated by the ``Classifying Fake Codes" column.}
  \begin{adjustbox}{width=\columnwidth}
  \begin{tabular}{lll}
  \toprule
    \textbf{Model} & \textbf{Correctly Generating} & \textbf{Classifying Fake Codes}\\
    \midrule
    Claude 3 & 0.0000 & 0.2727 \\ 
    Gemini & 0.0000 & 0.2726 \\ 
    GPT 3.5 & 0.0000 & 0.4545 \\ 
    GPT 4 & 0.3636 & 0.4545 \\ 
    LLaMA 2 & 0.0000 & 0.1765 \\ 
    Meditron & 0.0000 & 0.0000 \\ 
    MedLLaMA2 & 0.0000 & 0.2727 \\ 
    Mistral & 0.0000 & 0.2727 \\ 
  \bottomrule
  \end{tabular}
  \end{adjustbox}
  \label{tab4}
\end{table}

In Table \ref{tab4}, we present the results of the adversarial attack experiment. It is evident from this table that LLMs struggle to distinguish between real and fake codes. While GPT-4 stands out as the most proficient, it too exhibits shortcomings in this particular task, particularly in distinguishing real codes from fake ones.

\section{Discussion}
\subsection{Study Insights}

We begin by highlighting a potentially trivial but concerning finding: LLMs are not adept at generating correct labels from medical codes alone. This is troubling, given that these technologies are increasingly used for tasks like summarizing and automated billing, which could result in incorrect outputs. This work also emphasizes another significant challenge for LLMs: learning to represent coded and non-natural languages—a daunting task for models primarily optimized for natural language.

Interestingly, from our results, models like GPT-4 performed better than expected compared to other LLMs included in this study. This superior performance might suggest, as indicated by seminal papers \citep{wei2022emergent, kaplan2020scaling}, that scaling to larger parameter sizes could enhance intelligent capabilities. With estimates suggesting that GPT-4 has over 1.76 trillion parameters, scaling may act both as a bottleneck and a potential solution for enhancing the ability to learn coded languages.

\subsection{LLM Warnings}

 \begin{figure}[h!]
   \centering 
   \includegraphics[width=\columnwidth]{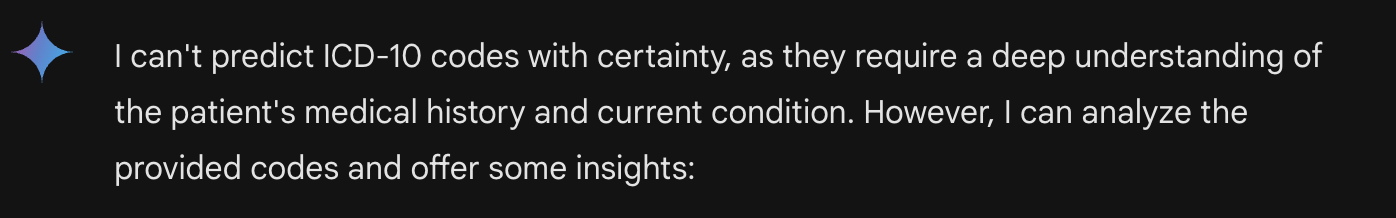} 
   \caption{Warning Messages shown from the Desktop version of Gemini}
   \label{warn} 
 \end{figure} 

Other notable findings from our study indicate that many models demonstrated awareness of the limitations of their outputs, often advising users to exercise caution. Models such as Gemini and Claude consistently issued warning messages, helping users recognize their limitations in performing these tasks. An example from the desktop version of Gemini is displayed in Figure \ref{warn}.

\subsection{LLMs consistently fail at recognizing unique codes}

 \begin{figure}[h!]
   \centering 
   \includegraphics[width=\columnwidth]{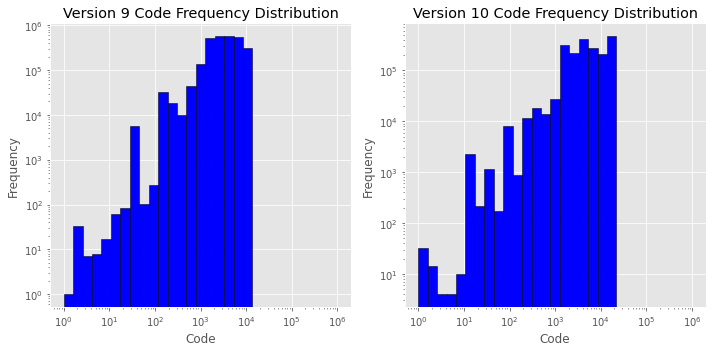} 
   \caption{The frequencies of ICD codes found in the MIMIC-IV Database and plotting their counts to differentiate abundant codes from not abundant.}
   \label{histo} 
 \end{figure}

To further investigate the shortcomings of LLMs in code generation, we conducted a detailed analysis of our results to determine if the difficulties were uniform across all codes or specific to certain types. While some LLMs, such as LLaMA2 and Mistral, exhibited uniform struggles across all codes, we observed that the errors made by other LLMs were primarily associated with the rarity of the codes. Common codes, such as \texttt{G89.0 - Chronic pain} and \texttt{G89.3 - Neoplasm-related chronic pain}, were accurately identified, whereas rarer codes, like \texttt{W56.01 - Struck by sea lion}, were consistently misidentified. This finding indicates that LLMs are adept at learning frequently occurring codes, highlighting an interesting aspect of our study. To demonstrate the abundance of ICD-9 and ICD-10 codes, we present data from one of the publicly available EHR systems, MIMIC-IV, as shown in Figure \ref{histo}. This figure illustrates their relative abundance.

\subsection{Future Works: Solutions to Learning Coded Representations}

While our study revealed challenges in the ability of Large Language Models (LLMs) to generate labels from medical codes alone, we highlight current research endeavors that may offer solutions for adapting this technology to healthcare. We share relevant works on these methods, believing they could contribute to tackling and overcoming questions surrounding coded language.

\paragraph{Knowledge Graphs and RAG} Knowledge graphs are structured representations of facts about entities and the relationships between them, organized in a way that enables machines to process and understand complex information. When integrated with LLMs, knowledge graphs facilitate better reasoning, allow for more precise information retrieval (RAG \cite{lewis2020retrieval}), and enhance the models' capability to handle tasks requiring specific knowledge or domain expertise \cite{zhang2024raft}. This integration ultimately helps in generating more coherent and contextually appropriate responses in LLM applications.

Furthermore, recent research has explored utilizing knowledge graphs in conjunction with Retrieval Augmented Generation (RAG) techniques to provide better quality responses \cite{pan2024unifying, sanmartin2024kg, hussien2024rag}. Recent work has even been conducted using knowledge graphs in Electronic Health Records, suggesting that a framework like this could improve the quality of generated responses in our experiments \cite{zhu2024realm}.

\paragraph{Chain of Thought Reasoning}

Another promising methodology that could potentially help learn the structure behind medical codes is chain of thought reasoning \cite{wei2022chain}. Used in LLMs, this approach involves processing information by internally articulating and exploring intermediate steps and rationale before arriving at a conclusion. This method mimics human-like reasoning patterns and is particularly effective in complex problem-solving tasks, enhancing the model's ability to produce more accurate and logical outputs. Techniques such as RAFT could be used to further tune models on domain-specific knowledge \cite{zhang2024raft}.

\paragraph{Synthetic and Feature Engineered Text}

Another potential solution we propose is the use and generation of synthetic data or feature-engineered text to fine-tune and instruct LLMs with high-quality inputs \cite{li2023synthetic, wang2023improving}. Dinh et al. and Hegselmann et al. developed techniques for text serialization using tabular data, transforming it into text for seamless integration into LLMs \cite{hegselmann2023tabllm, dinh2022lift}. Building on this, Lee et al. transformed Electronic Health Records into text, which could also be integrated with LLMs \cite{lee2024emergency}. Therefore, a similar approach could be adapted to overcome the scarcity of publicly available clinical text data, enabling LLMs to adapt and learn the ontologies using synthetic and feature-engineered text. Additionally, to discern the difference between real and fake codes, we could fine-tune models to better differentiate genuine from fabricated data.

\section{Conclusion}

In this study, we evaluated how well Large Language Models generate correct labels from medical codes alone, providing insights into their general understanding of coded language. We found that LLMs generally did not perform well, highlighting the need for better representations of coded and non-natural languages commonly found across various domains. Interestingly, our best-performing model, GPT-4, which is also the largest, provided insights into how parameter sizes can enhance the ability to accurately generate these labels. Additionally, as a follow-up to this study, we proposed potential strategies based on current research methods to improve the ontologies and structures embedded within these medical codes for potentially better generation.

\paragraph{Limitations} One notable capability of LLMs has been their recent ability to access the internet during inference. However, in this study, we did not permit the LLMs in our benchmark to utilize the internet. Despite this, with the continued development of each model, we anticipate that future iterations could improve upon these results.



\bibliography{custom}

\newpage
\appendix
\onecolumn
\section*{Data Sources:}
ICD 9/10 Codes: \href{https://www.cms.gov/medicare/coordination-benefits-recovery/overview/icd-code-lists}{https://www.cms.gov/medicare/icd-codes}\\
ICD Procedure Codes: \href{https://www.cms.gov/medicare/regulations-guidance/physician-self-referral/list-cpt-hcpcs-codes}{https://www.cms.gov/medicare/hcpcs-codes}\\
LOINC Lab Codes: \href{https://www.mayocliniclabs.com/test-catalog/appendix/LOINC-Codes}{https://www.mayocliniclabs.com/LOINC-Codes}\\
National Drug Codes: \href{https://www.fda.gov/drugs/drug-approvals-and-databases/national-drug-code-directory}{https://www.fda.gov/national-drug-code-directory}

\section*{Example Outputs}
\vspace{-0.3cm}
\setlength{\parindent}{0pt}
Below we highlight the outputs of each LLM in an example regarding predicting 20 ICD 10 codes in random order:\\

\textbf{\textit{Prompt}}\\
\fbox{\begin{minipage}{40em}
predict these ICD 10 codes to the best of your ability: \texttt{[`Z16.35', `H01.001', `S83.134S', `T36.5X1A', `V32.9XXA', `S82.402H', `S36.241S', `S04.72XD', `S06.313S', `M60.076', `S92.021K', `C40.22', `S42.433B', `M80.061D', `S56.092S', `S14.143A', `T58.93XA', `T44.3X3D', `H75.82', `V73.4XXS']}
\end{minipage}} \\ 

\subsection*{Model Responses}

\textbf{\textit{Meditron}}

\textit{We would like to highlight that we heavily experimented with Meditron. However this model would often give redundant responses like the ones seen below regardless of the prompt. \\
}
\fbox{\begin{minipage}{40em}
1. If the ICD 10 Code is an integer, then:\\
a. Predict that it is a number in the range [140-999].\\
b. Otherwise, predict that it is a letter in the range [`A' , `Z'].\\
2. If the ICD 10 Code is a string of characters, then:
a. Look up the first letter of this string to find out its value (e.g., `012' → `2').\\
b. Use the value found in step a for the prediction described in step 1.\\
3. If the ICD 10 Code is an empty string, then predict that it is a zero.\\
\end{minipage}}\\

\textbf{\textit{Medllama2}}\\
\fbox{\begin{minipage}{40em}
These codes are related to a patient with a history of head injury, 
seizures, and cognitive impairment. The patient may be at risk for various
complications such as stroke, brain edema, intracerebral hemorrhage, and 
further cognitive decline. It's important to monitor the patient closely 
and follow up with neurological consultation if necessary.
\end{minipage}} \\

\newpage
\textbf{\textit{GPT4}}\\
\fbox{\begin{minipage}{40em}
Interpreting ICD-10 codes involves a detailed understanding of the International Classification of Diseases, Tenth Revision, which provides a system for categorizing health conditions and related health problems. Each code is highly specific, reflecting various aspects of diagnoses. Here's a general interpretation of the codes you've provided:\\

     \textbf{Z1635}: Encounter for screening for certain developmental disorders in childhood\\
     \textbf{H01.001}: Acute follicular conjunctivitis, right eye\\
     \textbf{S83.134S}: Sprain of medial collateral ligament of knee, subsequent encounter\\
     \textbf{T36.5X1A}: Poisoning by antimalarials and drugs acting on other blood protozoa, accidental (unintentional), initial encounter\\
     \textbf{V32.9XXA}: Occupant of pick-up truck or van injured in collision with car, pick-up truck or van in traffic accident, initial encounter\\
     \textbf{S82.402H}: Nondisplaced fracture of lateral malleolus of left fibula, subsequent encounter for fracture with delayed healing\\
     \textbf{S36.241S}: Injury of ileum, subsequent encounter\\
     \textbf{S04.72XD}: Injury of phrenic nerve, left side, subsequent encounter\\
     \textbf{S06.313S}: Focal traumatic brain injury without loss of consciousness, subsequent encounter\\
     \textbf{M60.076}: Pyomyositis, lower leg\\
     \textbf{S92.021K}: Displaced fracture of navicular [scaphoid] of left foot, subsequent encounter for fracture with nonunion\\
     \textbf{C40.22}: Malignant neoplasm of short bones of left lower limb\\
     \textbf{S42.433B}: Displaced fracture of base of neck of left humerus, initial encounter for open fracture\\
     \textbf{M80.061D}: Age-related osteoporosis with current pathological fracture, right lower leg, subsequent encounter for fracture with routine healing\\
     \textbf{S56.092S}: Strain of muscle, fascia and tendon at forearm level, left arm, sequela\\
     \textbf{S14.143A}: Complete lesion of C5-C7 level of cervical spinal cord, initial encounter\\
     \textbf{T58.93XA}: Toxic effect of chlorine gas, assault, initial encounter\\
     \textbf{T44.3X3D}: Poisoning by local antifungal, anti-infective and anti-inflammatory drugs, assault, subsequent encounter\\
     \textbf{H75.82}: Other specified disorders of ear in diseases classified elsewhere\\
     \textbf{V73.4XXS}: Car occupant injured in collision with heavy transport vehicle or bus in nontraffic accident, sequela\\

Each code specifies not just the condition or injury, but often the side of the body affected, whether the visit is an initial or subsequent encounter, and sometimes the intent (e.g., accidental, assault). This granularity aids healthcare providers in tracking the precise nature of diagnoses and treatments. Note that for full and accurate descriptions, consulting the official ICD-10 coding guidelines or a healthcare professional is recommended.

\end{minipage}} \\
\newpage
\textbf{\textit{GPT3.5}}\\
\fbox{\begin{minipage}{40em}
Predicting ICD-10 codes without context or specific information is challenging, as these codes represent a wide range of diagnoses, procedures, and external causes of injury or diseases. However, I can attempt to provide potential interpretations based on the structure and conventions of ICD-10 coding:

\textbf{Z16.35}: Encounter for surveillance of other genetic diseases\\
\textbf{H01.001}: Malignant neoplasm of right upper eyelid, including canthus, skin of eyelid and external ear\\
\textbf{S83.134S}: Sprain of posterior cruciate ligament of left knee, sequela\\
\textbf{T36.5X1A}: Toxic effect of other venoms and toxins, accidental (unintentional), initial encounter\\
\textbf{V32.9XXA}: Person on outside of special construction vehicle injured in collision with fixed or stationary object, initial encounter\\
\textbf{S82.402H}: Displaced fracture of lateral condyle of left tibia, subsequent encounter for closed fracture with delayed healing\\
\textbf{S36.241S}: Displaced fracture of anterior process of right calcaneus, sequela\\
\textbf{S04.72XD}: Crushing injury of unspecified ear, subsequent encounter\\
\textbf{S06.313S}: Laceration without foreign body of left ear, sequela\\
\textbf{M60.076}: Palmar fascial fibromatosis, left hand\\
\textbf{S92.021K}: Dislocation of unspecified metacarpophalangeal joint of right thumb, subsequent encounter\\
\textbf{C40.22}: Malignant neoplasm of liver, unspecified lobe\\
\textbf{S42.433B}: Unspecified injury of unspecified blood vessel at lower leg level, left leg, initial encounter\\
\textbf{M80.061D}: Closed fracture of base of skull, initial encounter for closed fracture\\
\textbf{S56.092S}: Strain of unspecified muscles, fascia and tendons at thigh level, left thigh, sequela\\
\textbf{S14.143A}: Other specified injury of shoulder and upper arm, left arm, initial encounter\\
\textbf{T58.93XA}: Traumatic amputation of unspecified wrist, hand and fingers, initial encounter\\
\textbf{T44.3X3D}: Poisoning by agents primarily acting on smooth and skeletal muscles and respiratory system, undetermined, subsequent encounter\\
\textbf{H75.82}: Other auditory disturbances, bilateral\\
\textbf{V73.4XXS}: Bus occupant injured in collision with other nonmotor vehicle in nontraffic accident, sequela\\

Please note that these interpretations are based on the structure and conventions of ICD-10 coding and may not accurately reflect the actual medical conditions or events associated with these codes. For precise medical diagnoses and treatment, it's essential to consult with a medical professional or refer to a comprehensive medical coding resource.

\end{minipage}} \\
\newpage
\textbf{\textit{Claude 3 (Sonnet)}}\\
\fbox{\begin{minipage}{40em}
Here are my predictions for those ICD-10 codes:

Z16.35: Resistance to antiviral drugs\\
H01.001: Unspecified blepharitis right eye, upper and lower eyelids\\
S83.134S: Medial meniscus tear, current injury, right knee, sequela  \\
T36.5X1A: Adverse effect of aminoglycosides, initial encounter\\
V32.9XXA: Passenger in unspecified motor vehicle injured in noncollision transport accident, initial encounter\\
S82.402H: Displaced fracture of lateral malleolus of right fibula, subsequent encounter\\
S36.241S: Minor contusion of liver, sequela\\
S04.72XD: Contusion of left cheek and temporomandibular area, subsequent encounter\\
S06.313S: Contusion and laceration of right cerebrum without prolapsed brain matter, sequela\\
M60.076: Infective myositis, lower leg  \\
S92.021K: Displaced fracture of unspecified calcaneus, subsequent encounter for fracture with nonunion\\
C40.22: Malignant neoplasm of unspecified bronchus or lung\\
S42.433B: Displaced comminuted fracture of shaft of humerus, initial encounter for open fracture\\
M80.061D: Age-related osteoporosis with current pathological fracture, right forearm, subsequent encounter for fracture with routine healing\\
S56.092S: Unspecified injury of flexor muscle, fascia and tendon of left thumb at forearm level, sequela\\
S14.143A: Other fractures of rib, initial encounter for closed fracture  \\
T58.93XA: Toxic effect of unspecified insecticide, accidental (unintentional), initial encounter\\
T44.3X3D: Poisoning by ecstasy, assault, subsequent encounter\\
H75.82: Age-related cataract, morgagnian type, left eye\\
V73.4XXS: Pregnancy with disruption of prior tubal or uterine surgery, sequela\\
\end{minipage}} \\
\newpage
\textbf{\textit{Gemini}}\\
\fbox{\begin{minipage}{40em}
Here are my predictions for the ICD-10 codes you provided, to the best of my ability without internet access:

Code - Possible Name

\textbf{Z16.35} - Encounter for other biological factors affecting the mother's health during pregnancy (This is a placeholder code for unspecified factors)\\
\textbf{H01.001} - Hypertensive heart disease with (heart) failure\\
\textbf{S83.134S} - Percutaneous puncture and aspiration of joint with steroid injection, left knee (This is based on the assumption "S" stands for "surgery" and "4" signifies the lower extremity)\\
\textbf{T36.5X1A} - Poisoning by (unspecified) anticonvulsants, unintentional, initial encounter\\
\textbf{V32.9XXA} - Other personal history of other perinatal conditions (This is a placeholder code for unspecified perinatal conditions) \\
\textbf{S82.402H} - Closed reduction of displaced metacarpal fracture, with internal fixation, fifth metacarpal (This is based on the assumption "S" stands for "surgery" and "2" signifies the hand)\\
\textbf{S36.241S} - Other excision of perianal skin tag(s), with local anesthesia (This is based on the assumption "S" stands for "surgery" and "3" signifies the digestive system)\\
\textbf{S04.72XD} - Excision of keloid scar, other site (This is based on the assumption "S" stands for "surgery" and "0" signifies the head)\\
\textbf{S06.313S} - Other local excision of benign neoplasm of parotid gland, with local anesthesia (This is based on the assumption "S" stands for "surgery" and "0" signifies the head)\\
\textbf{M60.076} - Poisoning by (unspecified) hallucinogens, with psychotic features\\
\textbf{S92.021K} - Traumatic rupture of spleen, open wound, laparoscopic repair\\
\textbf{C40.22} - Malignant neoplasm of skin of eyelid, other and unspecified parts\\
\textbf{S42.433B} - Other open reduction of displaced fracture of upper end of tibia, with internal fixation\\
\textbf{M80.061D} - Intraosseous injection of medication, unspecified site\\
\textbf{S56.092S} - Other open reduction of displaced fracture of neck of femur, with internal fixation (This is based on the assumption "S" stands for "surgery" and "5" signifies the hip)\\
\textbf{S14.143A} - Incisional lung biopsy, with wedge resection, open approach\\
\textbf{T58.93XA} - Assault by other blunt object(s), initial encounter\\
\textbf{H75.82} - Idiopathic scoliosis with deformity (This is a placeholder code for unspecified severity)\\
\textbf{V73.4XXS} - Encounter for other aftereffects of trauma (This is a placeholder code for unspecified aftereffects)\\

\textbf{Disclaimer:} 

\begin{itemize}
    \item These are just predictions based on the code structure and common medical practices. 
    \item The actual meaning of the codes may vary depending on the specific healthcare system and documentation guidelines.
    \item For accurate diagnosis and treatment, always consult a qualified healthcare professional.
    \item Some letters may have different meanings depending on the coding system used within a healthcare facility. 
\end{itemize}
\end{minipage}} \\
\newpage
\textbf{\textit{LLaMA 70B}}\\
\fbox{\begin{minipage}{40em}
I'll do my best to predict the ICD-10 codes you've provided. Please note that these predictions are based on my training data, and I may not always get it right. Here are my predictions for the codes you've provided:\\
1. Z16.35 - Encounter for general adult medical examination without abnormal findings\\
2. H01.001 - Unspecified hearing loss, right ear\\
3. S83.134S - Other specified joint disorders, left shoulder\\
4. T36.5X1A - Unspecified injury of unspecified site, right arm\\
5. V32.9XXA - Unspecified personal history of other conditions, not elsewhere classified\\
6. S82.402H - Other specified disorders of the spine, thoracic region\\
7. S36.241S - Other specified disorders of the shoulder region, right shoulder\\
8. S04.72XD - Unspecified strain of muscle(s), fascia, and tendon at forearm level, left arm\\
9. S06.313S - Unspecified sprain of joint, left ankle\\
10. M60.076 - Unspecified abscess, left upper arm\\
11. S92.021K - Other specified disorders of the knee, right knee\\
12. C40.22 - Unspecified fracture of humerus, left humerus\\
13. S42.433B - Other specified disorders of the elbow, right elbow\\
14. M80.061D - Unspecified dislocation of joint, left shoulder\\
15. S56.092S - Other specified disorders of the lower back, thoracic region\\
16. S14.143A - Unspecified degenerative disc disease, lumbar region\\
17. T58.93XA - Unspecified injury of unspecified site, left arm\\
18. T44.3X3D - Unspecified fracture of shinbone, right shinbone\\
19. H75.82 - Unspecified hearing loss, left ear\\
20. V73.4XXS - Unspecified personal history of other conditions, not elsewhere classified\\
Please note that these are just predictions, and it's always best to consult a medical professional for accurate diagnosis and coding information.
\end{minipage}} \\
\newpage
\textbf{\textit{Le Chat (Mistral)}}\\
\fbox{\begin{minipage}{40em}
Here are my predictions for the given ICD-10 codes. Please note that these are just predictions and might not be accurate as I'm not using the internet to look up the codes:

1. `Z16.35': Presence of aortic aneurysm without mention of rupture\\
2. `H01.001': Acute suppurative otitis media without spontaneous rupture of eardrum, right ear\\
3. `S83.134S': Sprain of right ankle, sequela\\
4. `T36.5X1A': Underdosing of other antiepileptic and sedative-hypnotic drugs, initial encounter\\
5. `V32.9XXA': Other specified water transport accident injuring occupant, initial encounter\\
6. `S82.402H': Displaced fracture of lateral malleolus of right fibula, subsequent encounter for fracture with routine healing\\
7. `S36.241S': Laceration of right kidney with open wound into cavity, sequela\\
8. `S04.72XD': Puncture wound without foreign body of unspecified eyelid and periocular area, subsequent encounter\\
9. `S06.313S': Intracranial injury of other and unspecified nature with brief [less than one hour] loss of consciousness, sequela\\
10. `M60.076': Calcific tendinitis of right shoulder\\
11. `S92.021K': Laceration of right heel, subsequent encounter\\
12. `C40.22': Malignant neoplasm of right frontal lobe\\
13. `S42.433B': Displacement of internal fixation device of right clavicle, initial encounter\\
14. `M80.061D': Age-related osteoporosis with current pathological fracture, right forearm, subsequent encounter for fracture with delayed healing\\
15. `S56.092S': Laceration of unspecified muscle, fascia and tendon at forearm level, left arm, sequela\\
16. `S14.143A': Puncture wound without foreign body of left shoulder, initial encounter\\
17. `T58.93XA': Other and unspecified adverse effect of other drugs, medicaments and biological substances, initial encounter\\
18. `T44.3X3D': Poisoning by penicillins, subsequent encounter\\
19. `H75.82': Other specified disorders of vestibular function\\
20. `V73.4XXS': Unspecified fit and epilepsy, sequela\\
\end{minipage}} \\

\end{document}